\documentclass[10pt,twocolumn,letterpaper]{article}

\usepackage[pagenumbers]{cvpr} 

\usepackage{multirow}
\usepackage{makecell} 
\usepackage{animate}
\usepackage{graphicx}
\usepackage{xcolor}
\usepackage[misc]{ifsym}
\usepackage{cuted}

\usepackage[utf8]{inputenc} 
\usepackage[T1]{fontenc}    
\usepackage{booktabs}       
\usepackage{amsfonts}       
\usepackage{nicefrac}       
\usepackage{microtype}      
\usepackage{amsmath}

\usepackage{threeparttable}
\usepackage{wrapfig}
\usepackage{caption}
\usepackage{subcaption}
\usepackage{siunitx}
\definecolor{cvprblue}{rgb}{0.21,0.49,0.74}
\usepackage[pagebackref,breaklinks,colorlinks,allcolors=cvprblue]{hyperref}
\usepackage{makecell}

\usepackage[capitalize]{cleveref}
\crefname{section}{Sec.}{Secs.}
\Crefname{section}{Section}{Sections}
\Crefname{table}{Table}{Tables}
\crefname{table}{Tab.}{Tabs.}

\newcommand{\name}{\textsc{Talk2Move}}

\newcommand\blfootnote[1]{%
  \begingroup
  \renewcommand\thefootnote{}\footnote{#1}%
  \addtocounter{footnote}{-1}%
  \endgroup
}

\title{\name: Reinforcement Learning for Text-Instructed  Object-Level  Geometric Transformation in Scenes}

\author{%
Jing Tan$^{1,4}$\footnotemark 
\quad
Zhaoyang Zhang$^{1}$ \quad
Yantao Shen$^{1}$ \quad
Jiarui Cai$^{2}$ \\
Shuo Yang$^{1}$ \quad
Jiajun Wu$^{3}$ \quad
Wei Xia$^{1}$ \quad
Zhuowen Tu$^{1}$ \quad
Stefano Soatto$^{1}$ \vspace{0.2em}\\
$^{1}$AWS Agentic AI \quad $^{2}$Amazon Web Services \quad  $^{3}$Amazon Robotics \quad  $^{4}$CUHK
}

\begin{document}

\twocolumn[{
\renewcommand\twocolumn[1][]{#1}
\maketitle
\begin{center}
    \centering
    \includegraphics[width=\linewidth]{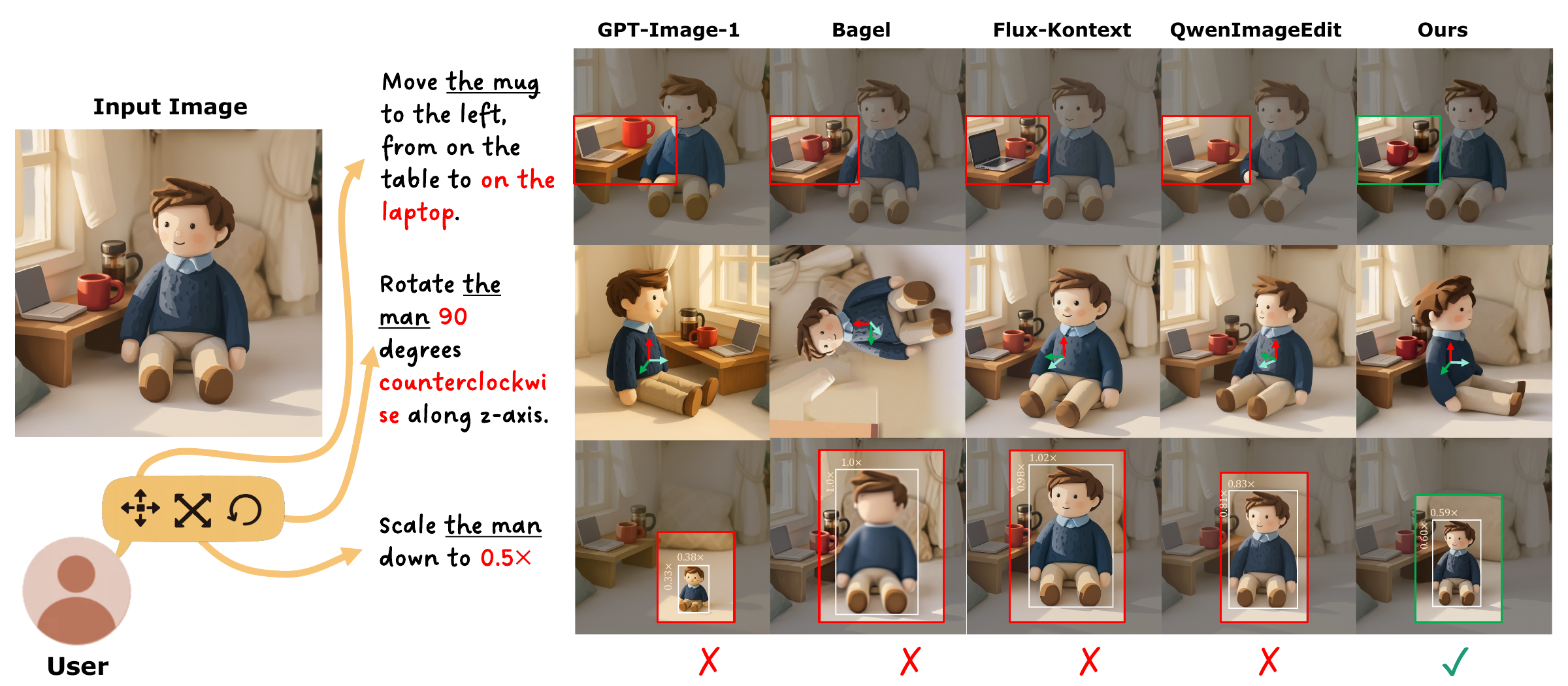}
    \centering
    \captionof{figure}{\small We introduce \textbf{\name}, a text-guided scene editing model for object-level geometric transformation, focusing on object translation, rotation and resizing, achieving superior results over current SOTA image editing models. 
    }
	\label{fig:teaser}
\end{center}
}]

\blfootnote{\small $^*$ The work was done during an internship at AWS Agentic AI. Jing Tan is now with the Chinese University of Hong Kong. Correspondence to ozhaozha@amazon.com}

\begin{abstract}
We introduce Talk2Move, a reinforcement learning (RL) based diffusion framework for text-instructed spatial transformation of objects within scenes. Spatially manipulating objects in a scene through natural language poses a challenge for multimodal generation systems. While existing text-based manipulation methods can adjust appearance or style, they struggle to perform object-level geometric transformations—such as translating, rotating, or resizing objects—due to scarce paired supervision and pixel-level optimization limits. Talk2Move employs Group Relative Policy Optimization (GRPO) to explore geometric actions through diverse rollouts generated from input images and lightweight textual variations, removing the need for costly paired data. A spatial reward guided model aligns geometric transformations with linguistic description, while off-policy step evaluation and active step sampling improve learning efficiency by focusing on informative transformation stages. Furthermore, we design object-centric spatial rewards that evaluate displacement, rotation, and scaling behaviors directly, enabling interpretable and coherent transformations.
Experiments on curated benchmarks demonstrate that Talk2Move achieves precise, consistent, and semantically faithful object transformations, outperforming existing text-guided editing approaches in both spatial accuracy and scene coherence.
\end{abstract}  
\section{Introduction}
\label{sec:intro}
Object transformation is essential in scene image editing, enabling precise control over spatial layout and object interactions. These geometric manipulations let users adjust content to match their intent, such as \textit{“move the mug to the left”} or \textit{“rotate the man 90 degrees counterclockwise.”} Guiding these transformations through natural-language instructions is particularly valuable, as text provides an intuitive and low-barrier interface that makes spatial editing accessible to a broad range of users.

Current graphics-based editing methods~\cite{dragyourgan,mou2023dragondiffusion,3ditscene,li2024omnidrag} often rely on 2D/3D primitives as control signals to assist scene editing. 
This requires human intervention and domain expertise in 2D/3D generation, making it {\bf not user-friendly} nor in natural form of interaction for general users. 
Moreover, these approaches typically {\bf demands large amounts of high-quality paired data} for SFT training, yet such data are rare and expensive to obtain. 
Spatial-manipulation pairs are particularly scarce—video datasets or 3D simulations provide only limited examples and are costly to scale. 
Top-down modeling of image content exists \cite{gallagher2015happened,reed2015deep} which requires small training data but their performances are not meeting the practical requirement.
Moreover, SFT-based methods rely on pixel-level MSE loss that struggles to disentangle objects from scenes, leading to limited spatial control.

In this work, we introduce~\name, the first RL-based framework that leverages \textbf{text guidance} to perform geometric object-level transformations for scene editing.
Built upon a flow-based GRPO paradigm, \name~perturbs diffusion trajectories through stochastic noise injection to explore diverse spatial transformations during training. 
It scales efficiently by generating diverse rollouts with simple prompt variations, eliminating the need for costly paired data.
To disentangle objects from scene backgrounds,~\name~employs a spatially grounded reward model that directly evaluates object displacement, rotation, and scaling behaviors beyond pixel-level similarity, leading to interpretable and geometry-aware optimization objectives.

However, conventional dense rollouts in GRPO training are computationally expensive, and not all denoising steps contribute equally to learning.
Therefore, we introduce a step-importance measure that quantifies how each perturbed denoising step influences the task reward.
Building on this, we propose step-wise active sampling, which adaptively selects informative denoising steps through lightweight off-policy reward evaluation and skips redundant ones via skip connections.
This design substantially improves training efficiency by 2$\times$ while preserving the reward robustness.

To summarize, our contributions are four-fold.
\begin{itemize}
\itemsep 0em 
\item To the best of our knowledge, we are the first to formulate the text-guided geometric object transformation problem within a reinforcement learning framework, enabling the editing model to follow textual instructions for precise spatial manipulation.
\item Our pipeline is data-efficient compared to SFT-based methods, reducing dependence on costly paired annotations. We design a spatial-aware reward model that separates target objects from the scene to improve geometric transformation learning, and introduce an early-exit mechanism for flow-based GRPO to focus training on the most informative steps and lower sampling cost.
\item We establish an effective data collection pipeline that provides high-quality data for geometric object transformations, and we demonstrate that this pipeline can be scaled to produce large quantities of input pairs suitable for GRPO online training.
\item Experiments demonstrate that~\name~significantly improves spatial accuracy and scene coherence under limited budgets, achieving state-of-the-art spatial transformation performance across both open-source and proprietary editing systems. 
\end{itemize}

\section{Related Work}
\noindent\textbf{Drag-based Spatial Manipulation.} Drag-based editing enables explicit object control through handle–target interactions. 2D methods~\cite{dragyourgan,mou2023dragondiffusion,dragyournoise,shi2024dragdiffusion,shin2024instantdrag,lu2024regiondrag} rely on point tracking or feature-space guidance to iteratively move objects, but they require manual specification of control points and struggle with complex, high-level instructions.

To introduce geometric priors, 3D-aware approaches~\cite{3dfixup,3ditscene,imagesculpting} lift objects into 3D space and perform transformations on reconstructed 3D representation. However, these pipelines involve multi-stage 2D-to-3D lifting and rendering, leading to high complexity, cumulative errors, and reduced image fidelity. Overall, both 2D and 3D dragging methods depend on explicit spatial primitives and user intervention, making them less suitable for natural scene-level object manipulation.

\noindent\textbf{Text-based Image Editing.}
Recent image generation and editing methods fall broadly into two families: diffusion-based models and LLM/VLM-connected frameworks. Diffusion-based methods~\cite{stablediffusion,ramesh2021zero} generate images via denoising, with recent flow-matching models such as Flux and Flux.1 Kontext~\cite{fluxkontext,blackforestlabs2023flux} improving convergence and edit consistency through in-context learning. However, relying solely on text tokenizers limits their ability to follow fine-grained spatial instructions.

Another line of work integrates pre-trained LLMs/VLMs with diffusion decoders via lightweight adapters. Methods such as MetaQueries~\cite{metaqueries}, Seed-X~\cite{seedx}, and Emu2~\cite{emu2}, as well as unified transformer frameworks like MoT and Bagel~\cite{mot,bagel}, enhance prompt following but often struggle to preserve visual coherence when only ViT-based encoders are used. QwenImageEdit~\cite{qwenimage} addresses this by combining VLM semantic features with VAE reconstructive features for stronger alignment between instruction following and visual fidelity. We adopt QwenImageEdit as our backbone for spatial manipulation tasks.

\noindent\textbf{Reinforcement Learning for Visual Generation.}
Reinforcement learning for diffusion-based generation typically formulates the denoising process as a Markov Decision Process (MDP). Reward-Weighted Regression~\cite{rwr} models generation as a one-step MDP and shifts the image distribution toward high-reward samples. DDPO~\cite{ddpo} extends this to a multi-step MDP by treating each denoising step as an action, making it well-suited for deterministic samplers without requiring repeated responses from the same query.
Recent works adapt GRPO to diffusion models. FlowGRPO~\cite{flowgrpo} and DanceGRPO~\cite{dancegrpo} introduce stochastic drift into DDPM and Rectified Flow samplers, enabling multiple rollout paths around the deterministic trajectory. Subsequent improvements~\cite{mixgrpo,branchgrpo,g2rpo} reduce training cost via sliding-window optimization or tree-based trajectory reorganization with reward backpropagation. Although effective, these methods remain computationally expensive during sampling.
In this work, we further improve time efficiency by introducing an early-exit mechanism to accelerate rollout generation. 
TempFlow-GRPO~\cite{he2025tempflow} shows that temporal reweighting improves general GRPO stability. Our work instead focuses on identifying task-dependent transformation stages for object-level geometry.
RL-based image editing methods \cite{li2024instructrl4pix,editscore,li2025uniworld} exist but they are not focused on the object transformation tasks.

\begin{figure*}[ht]
    \centering
    \includegraphics[width=\linewidth]{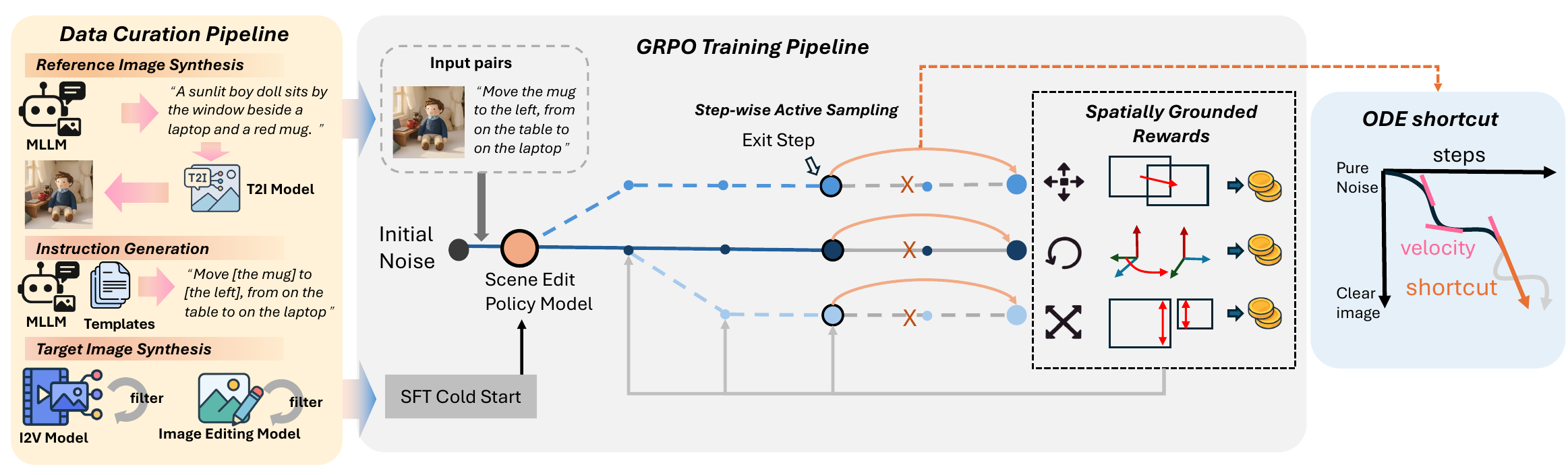}
    \caption{\small \textbf{The pipeline of~\name.}~\name~streamlines a GRPO-style reinforcement learning pipeline tailored for flow-based image editing. Starting from an initial noise sample, stochastic perturbations are injected at each diffusion step to generate diverse sampling trajectories.
Spatially grounded rewards from specialist models, which explicitly evaluate object-level geometric changes, are then used to compute group-relative advantages for policy gradient updates.}
    \label{fig:pipeline}
    \vspace{-8pt}
\end{figure*}

\section{Geometric Transformation for Object-Level Scene Editing}
\noindent\textbf{Problem Formulation.~}
We study text-guided geometric transformation for object-level scene editing, where the goal is to modify an object’s position, orientation, and scale in an image according to a text instruction, while keeping the scene static and unedited regions consistent. We focus on three fundamental spatial operations: object translation, rotation, and resizing, which together span the core dimensions of geometric object transformation.

Formally, we define a spatial editing model as a mapping $f_\theta : (I, T) \rightarrow I'$, where $I$ is the input scene image, $T$ is a text instruction, and $I'$ is the edited output. To regularize the instruction–action space, we use a set of predefined transformation templates ${t_k}$ for translation, rotation, and resizing. Each template specifies a target object and a spatial relation (e.g., “move the mug to the left”), providing a consistent basis for supervision and evaluation.

\noindent\textbf{Standardized Transformation Templates.~}
For object translation, the template defines both the movement direction (left, right, up, down, forward, backward) and the target spatial reference (e.g., ``on top of the sofa'', ``near the violin''). 
For object rotation, it specifies the rotation axis ($x$, $y$, $z$), direction (clockwise or counterclockwise), and angle (45$^\circ$, 90$^\circ$, 135$^\circ$, 180$^\circ$). 
The rotation axes are defined in the object's local coordinate frame, which aligns with the object's inherent orientation in the scene. 
For object resizing, the template defines the scaling ratio (e.g., 1.25$\times$, 1.5$\times$, 2$\times$, 3$\times$, 4$\times$). 
Together, these standardized templates constrain the open object transformation problem into a well-defined and verifiable space suitable for reproducible learning.
\section{Geometric Transformation Data Curation}
GRPO-style RL training requires only input samples consisting of a reference image and a text prompt. 
We take the object translation task as an example to illustrate the data curation process. 

\noindent\textbf{Reference Image Generation.}
We prompt a large language model (LLM) to generate textual scene descriptions that specify realistic scenes with a background and multiple sparsely distributed movable objects, and then employ open-source text-to-image models~\cite{fluxkontext,qwenimage} to synthesize the corresponding reference images.

\noindent\textbf{Instruction Generation.~} 
For each generated image, we use a vision-language model (VLM) to annotate spatial editing instructions with predefined templates. 
These annotations are parsed into a structured format that preserves geometric attributes such as object location, movement direction, rotation angle, and resizing ratio for reward computation. 
Given a reference image, we can expand the dataset by varying textual prompts to generate many training samples and extend this process to other object transformation tasks, substantially reducing data creation costs compared to paired supervision. In total, we construct 3,200 samples over 800 unique images for object translation. We leave large-scale data expansion to future work, as the current dataset is sufficient for training.

\noindent\textbf{Target Image Synthesis for SFT Cold Start.}
Current image editing backbones struggle with precise spatial manipulation, so we first apply an SFT cold-start to learn basic spatial-editing priors from a small amount of high-quality ground-truth data. However, constructing paired spatial-editing data is challenging, as real-world datasets rarely provide matched before–after examples for specific geometric transformations. To address this, we synthesize supervision with tailored strategies for translation, rotation, and resizing.
For object translation and rotation, we use the API-based video generation model~\cite{gao2025seedream} to simulate physically plausible object motion in diverse scenes. Conditioned on a reference image as the first frame and a manipulation prompt, the model generates short videos depicting the desired transformation. After filtering by spatial accuracy and visual consistency, we obtain 800 valid translation pairs and 43 high-quality rotation pairs.
For object resizing, where current video generation models perform poorly, we instead rely on open-source image editing models~\cite{qwenimage,fluxkontext} to synthesize coarse up/down scaling. The edited images are paired with their references and filtered via perceptual matching, yielding 110 diverse resizing pairs.

\section{Method}
\subsection{Preliminary}
Flow-based GRPO~\cite{flowgrpo,dancegrpo} extends GRPO to flow-based generation by modeling the denoising trajectory as a Markov Decision Process (MDP). 
During inference, the diffusion model usually starts from a random Gaussian noise $\mathbf{x}_T$ and performs a reverse denoising process over $T$ steps. At each step, the model predicts the noise component conditioned on the input signals (e.g., text or image features) and progressively removes it, gradually reconstructing a clean image $\mathbf{x}_0$.
Each denoising step can be viewed as a Reinforcement Learning (RL) state $s_t = (\mathbf{c}, \mathbf{x}_t)$, where $\mathbf{x}_t$ is the noisy latent and $\mathbf{c}$ is the conditioning input. The action corresponds to predicting the previous latent $a_t = \mathbf{x}_{t-1}$, and the policy is defined as the conditional generation step:
$
\pi(a_t | s_t) = p_\theta(\mathbf{x}_{t-1} | \mathbf{x}_t, \mathbf{c}).
$
The entire $T$-step denoising trajectory thus forms an MDP episode. At each step, the log probability represents the likelihood of sampling the previous latent $\mathbf{x}_{t-1}$ under the model-predicted conditional Gaussian transition $p_\theta(\mathbf{x}_{t-1} | \mathbf{x}_t, \mathbf{c})$.

Following this formulation, the GRPO objective can be extended to diffusion-based models as:
\begin{equation}
\small
\begin{aligned}
J_{\mathrm{GRPO}}(\theta) &= 
 \mathbb{E}_{\mathbf{c}, \{x_i\} \sim \pi_{\text{old}}(\cdot | \mathbf{c})} 
\Bigg[
\frac{1}{G \cdot T} 
\sum_{i, t} 
\min\big( 
r_t^i(\theta) \hat{A}_t^i, \\[-2pt]
& \mathrm{clip}\left(r_t^i(\theta), 1 - \epsilon, 1 + \epsilon\right) \hat{A}_t^i 
\Bigg],
\end{aligned}
\label{eq:flowgrpo}
\end{equation}
where
$r_t^i(\theta) = \frac{p_\theta(\mathbf{x}_{t-1} | \mathbf{x}_t, \mathbf{c})}{p_{\text{old}}(\mathbf{x}_{t-1} | \mathbf{x}_t, \mathbf{c})},$
$G$ denotes the number of stochastic rollouts per input.

Flow-based models typically solve Ordinary Differential Equations (ODE) at each denoising step and produce deterministic results for each input. Therefore, flow-based GRPO methods inject stochasticity in the sampling direction by adding noise perturbation to each ODE step, turning it into an analogous to a Stochastic Differential Equation (SDE) formulation to obtain a group of rollouts.
\subsection{Training Pipeline}
In this section, we describe the overall training pipeline of~\name. We first use an SFT cold start to equip the editing model with basic geometric transformation ability, and then present the RL stage in detail.

\noindent\textbf{Cold Start.}
We perform lightweight LoRA-based fine-tuning to embed basic spatial-editing capabilities into the diffusion backbone with a small amount of groundtruth data. The annotated image pairs from all three tasks are merged together and train a unified SFT checkpoint. LoRA adapters are inserted into attention projection layers, normalization layers, and linear layers, while the text encoder, VAE, and ViT backbones are frozen. 
This initialization provides the model with coarse spatial priors, improving rollout stability and sample quality during GRPO training.

\noindent\textbf{RL implementation.}
To achieve data-efficient and spatially accurate scene manipulation, \name~adopts a GRPO-style reinforcement learning framework tailored for diffusion policies. Starting from an initial noise sample, we follow the flow-based GRPO paradigm~\cite{flowgrpo, dancegrpo}, injecting stochastic perturbations at each diffusion step to form groups of diverse sampling trajectories. These rollouts capture a rich set of spatial transformations conditioned on the input prompt and diffusion dynamics. Spatially grounded rewards are provided by off-the-shelf specialist models that disentangles objects explicitly from the scene and evaluates object-level transformations, from which we compute group-relative advantages to update model parameters via policy gradients.

\subsection{Efficient GRPO with Early Exit}
The dense rollout generation in GRPO introduces substantial computational overhead. Empirically, we observe that not all denoising steps contribute equally. Perturbations at some steps even degrade sampling performance. This motivates the need to identify and prioritize steps that provide stronger learning signals.

\noindent\textbf{Off-policy Step Evaluation.}
Inspired by reasoning LLMs~\cite{wang2025beyond}, which emphasize tokens with high intrinsic uncertainty, we propose to focus on diffusion steps exhibiting high \textit{extrinsic} uncertainty. Since the noise variance in diffusion is fixed by the scheduler, we measure uncertainty through the variance of rollout rewards at each step. High reward variance indicates strong exploration potential and thus higher learning utility, analogous to high-entropy tokens in language RL.
In addition, according to MixGRPO~\cite{mixgrpo} and our empirical findings, early denoising steps largely determine global layout, while later steps refine fine-grained details. Therefore, the step importance is defined as the reward variance of rollouts when exiting from each denoising step. 
In practice, starting from step 0, we incrementally perturb small sets of denoising steps on a few input images (2--4) and compute rollout rewards. This efficient calibration, requiring only a single GPU, yields a task-specific, step-aware reward distribution. The optimal exit step is then selected as the last perturbed step achieving the maximum reward variance.

\noindent\textbf{Active Step Sampling.}
With the exit step identified, we introduce \textit{step-wise active step sampling}, which takes ODE shortcuts at the exit step to bypass redundant later steps, thereby improving both rollout and training efficiency.

\begin{figure}[t]
    \centering
    \includegraphics[width=\linewidth]{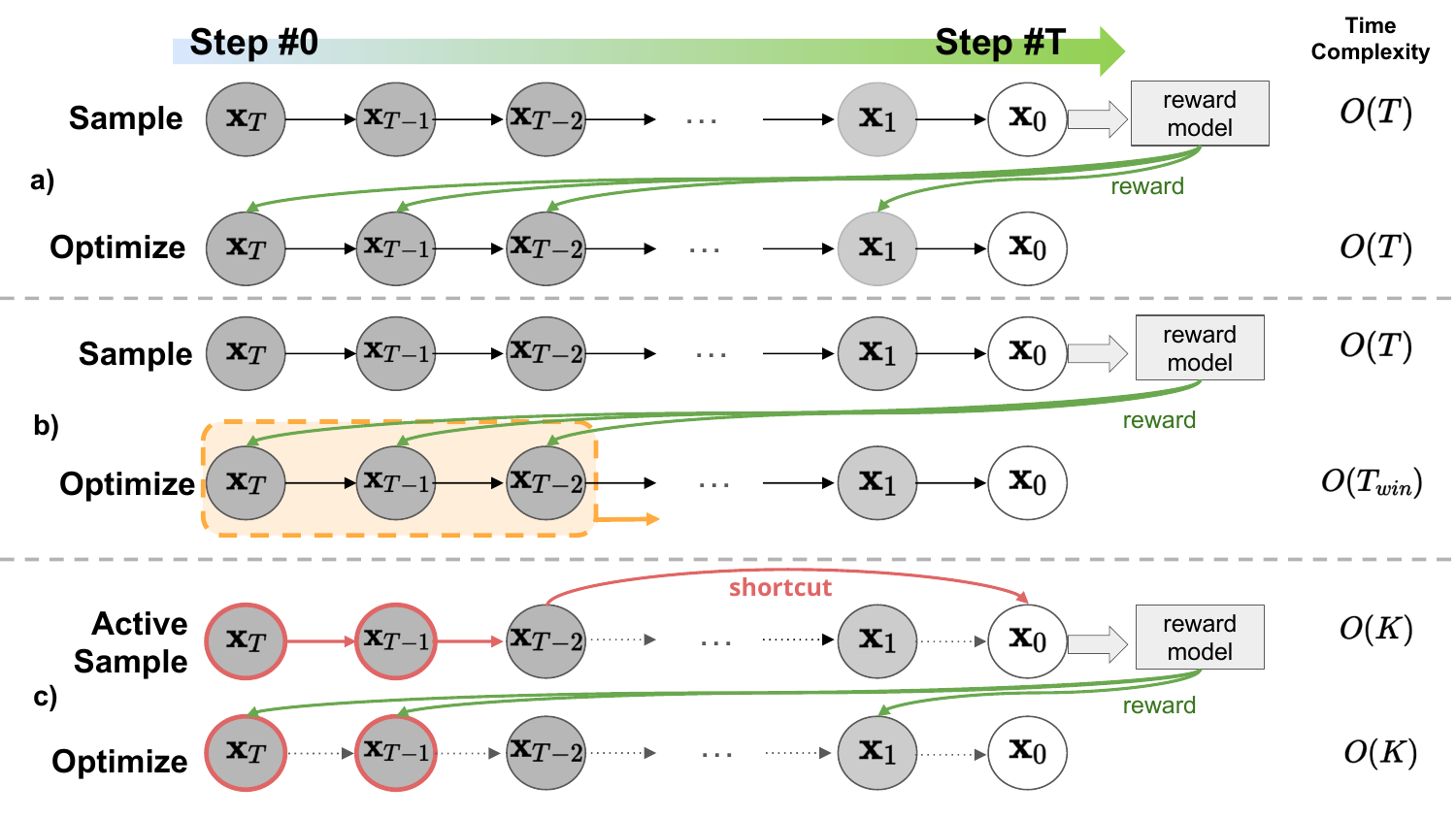}
    \caption{\small \textbf{Three types of step sampling}: \textbf{(a)} is the full sampling and optimizing GRPO~\cite{flowgrpo,dancegrpo}; subsequent methods~\cite{mixgrpo,branchgrpo} as in \textbf{(b)}, use a sliding window (yellow) to reduce the optimizing steps per iteration; \textbf{(c)} our work introduces step-wise active sampling that select the informative steps (red) and use shortcuts to bypass the rest of the steps, reducing both the sampling and optimizing time.}
    \vspace{-14pt}
    \label{fig:active_sampling}
\end{figure}

Flow-based GRPO approaches apply ODE updates at every denoising step, resulting in a time complexity of $T(t_{\text{sample}} + t_{\text{optim}})$ for $T$ steps. The sliding-window mechanism in~\cite{mixgrpo} alleviates this by optimizing only a subset of steps per iteration. We further enhance efficiency by directly denoising from the exit step $K$ to the final step $T$ using model predictions (see \cref{fig:active_sampling}), effectively shortening the trajectory. This reduces total computation to approximately $K(t_{\text{sample}} + t_{\text{optim}})$, where $K \leq T$. As long as $K < T$, our method achieves consistently higher training efficiency while maintaining comparable task performance.

\subsection{Spatially Grounded Reward Design}

The choice of reward model is one of the most important factors in GRPO training, as it provides critical information for the policy update direction. 
Previous works~\cite{flowgrpo,editscore} employ holistic, image-level metrics such as aesthetic rewards, CLIP-based alignment, or VLMs fine-tuned for common editing tasks.
These rewards serve as a coarse measure of overall image quality or human preference. 
We extend the reward model to fine-grained, spatially grounded rewards tailored for spatial manipulation editing.

To evaluate the correctness of geometric object transformation, we first localize the prompted object in both the reference and edited images using text-driven segmentation~\cite{sam,groundingdino} to obtain segmentation masks and bounding boxes. 
Each manipulation task defines a specific spatial reward. 
For object translation, we compute the relative displacement between the reference and edited object centers following the GenEval~\cite{geneval} protocol. We also employ depth estimation model~\cite{depthanything,wang2025moge} to obtain object depth value before and after editing to reward forward and backward movements.
For rotation, we use Orient-Anything~\cite{orientanything} to estimate orientations for object crops from the scene image and evaluate alignment along the prompted axis, direction, and angle. 
For resizing, we compare the normalized size ratio differences of reference and edited object bounding boxes. 
Although each reward differs in formulation, all share a unified goal, which is to measure the alignment between the prompted and achieved spatial transformations in normalized coordinate space. 
These task-specific rewards provide fine-grained feedback to guide policy updates toward spatially accurate editing behavior.
\section{Experiments}
\begin{table*}[t]
    \small
    \centering
    \setlength{\tabcolsep}{5pt}
    \renewcommand{\arraystretch}{0.8} 
        \caption{\small \textbf{Quantitative comparison} of object transformation tasks on our curated \textbf{synthetic} test benchmark in terms of editing accuracy, average translation distance and editing errors. We also report \textbf{human evaluation} results in terms of winning rate. We note that win rate does not capture second-best or near-tie preferences, and thus may underestimate methods that produce competitive but less visually salient results, such as GPT-based models.} 

\scalebox{0.95}{
\begin{tabular}{@{}lccccccccc@{}}
\toprule
\multicolumn{1}{c}{\multirow{4}{*}{Methods}} &
\multicolumn{3}{c}{Translation} &
\multicolumn{3}{c}{Rotation} &
\multicolumn{3}{c}{Resize} \\ 
\cmidrule(lr){2-4} \cmidrule(lr){5-7} \cmidrule(lr){8-10}
\multicolumn{1}{c}{} &
Trans. Dist. $\uparrow$ & Acc. $\uparrow$ & \makecell{Human \\ Win Rate $\uparrow$} &
Rot. Err. $\downarrow$ & Acc. $\uparrow$ & \makecell{Human \\ Win Rate $\uparrow$} &
Scale Err. $\downarrow$ & Acc. $\uparrow$ & \makecell{Human \\ Win Rate $\uparrow$} \\ 
\midrule
GPT-Image-1~\cite{openai2024gptimage1} & 0.5416 & 64.29\% & 26.25\% & 0.4293 & 2.33\% & 6.25\% & \textbf{0.3501} & 5.08\% & 1.39\% \\
Flux-Kontext~\cite{fluxkontext} & 0.0499 & 4.41\% & 1.25\% & 0.4259 & 6.82\% & 16.25\% & 0.4318 & 1.67\% & 15.28\% \\
Bagel~\cite{bagel} & 0.1705 & 14.49\% & 2.50\% & 0.3240 & 13.64\% & 1.25\% & 0.4790 & 0.00\% & 8.33\% \\
QwenImageEdit~\cite{qwenimage} & 0.2551 & 32.86\% & 12.50\% & 0.4129 & 9.30\% & 7.50\% & 0.4203 & 7.50\% & 11.11\% \\
\textbf{Ours} & \textbf{0.6667} & \textbf{76.67\%} & \textbf{ 57.50\% } & \textbf{0.2861} & \textbf{29.55\%} & \textbf{68.75\%} & 0.3894 & \textbf{9.17\%} & \textbf{63.89\%} \\ 
\bottomrule
\end{tabular}

}
    \vspace{-4pt}
  \label{tab:quantitative_translation}
\end{table*}

\begin{table}[t]
\centering
\small
\setlength{\tabcolsep}{0pt}
\renewcommand{\arraystretch}{1.0}
\caption{\small \textbf{Quantitative comparison} of object transformation tasks on curated \textbf{real} images from OpenImages-V6~\cite{OpenImages2} in terms of editing accuracy, average translation distance and editing errors. } 
\scalebox{0.83}{\begin{tabular}{lcccccc}
\toprule
\multirow{2}{*}{{Methods}} &
\multicolumn{2}{c}{{Translation}} &
\multicolumn{2}{c}{{Rotation}} &
\multicolumn{2}{c}{{Resize}} 
\\ \cmidrule(lr){2-3}\cmidrule(lr){4-5}\cmidrule(lr){6-7}
& Trans. Dist.$\uparrow$ & Acc.$\uparrow$ 
& Rot. Err.$\downarrow$ & Acc.$\uparrow$ 
& Scale Err.$\downarrow$ & Acc.$\uparrow$
\\
\midrule
GPT-Image-1~\cite{openai2024gptimage1} & 0.3237 & 23.08\% & \textbf{0.1774} & 25.00\% & 0.6898 & \textbf{7.14\%} \\
Flux-Kontext~\cite{fluxkontext} & 0.1992 & 26.92\% & 0.2567 & 6.67\% & 0.7120 & 1.79\% \\
Bagel~\cite{bagel} & {0.2258} & 15.38\% & 0.3700 & 6.25\% & {0.7159} & 3.57\% \\
QwenImageEdit~\cite{qwenimage} & \underline{0.3725} & \underline{42.31\%} & 0.2101 & 25.00\% & \underline{0.6048} & 1.79\% \\
\textbf{Ours} & \textbf{0.5196} & \textbf{53.85\%} & \underline{0.1997} & \textbf{31.25\%} & \textbf{0.5947} & \textbf{7.14\%} \\
\bottomrule
\end{tabular}}
\vspace{-6pt}
\label{tab:quantitative_real}
\end{table}

\subsection{Implementation Details}
For cold start, we train Qwen-Image-Edit~\cite{qwenimage} with rank-64 LoRA layers for 3,000 iterations and with a learning rate of 1e-4.
For GRPO training, we build upon the FlowGRPO~\cite{flowgrpo} baseline, with sample noise level of 1.0 and a clip range of 2e-4, on 16 H200 GPU server. 
The total training time for each subtask is $\sim 160$ GPU hours.

\subsection{Evaluation Metrics}
To evaluate the spatial accuracy for scene editing, we introduce task-specific metrics. 

\noindent\textbf{Translation.} We report the Translation Distance (Trans. Dist.) that measures the relative distance between object center points in the reference and edited image; and the translation Accuracy (Acc.) that measures the success rate. An edit is considered successful only if it passes all four criteria: 1) correct movement: following GenEval~\cite{geneval}, we verify whether the detected object in the edited image has moved in the specified direction relative to its position in the reference image; 2) object identity preservation: the edited object must remain the same object (CLIP similarity > 0.75); 3) no duplication: the object should no longer appear at its original location (similarity < 0.95); 4) scene preservation: the overall scene should remain consistent (background L1 distance $\leq 0.2$).

\noindent\textbf{Rotation.} We compute rotation accuracy by verifying whether the edited rotation angle lies within ±\ang{20} of the target angle. We also report rotation error, defined as the average normalized difference between the target (prompted) angle and the predicted (edited) angle.

\noindent\textbf{Resizing.}
We measure resizing accuracy by checking whether the edited size falls within a ±10\% tolerance of the target scaling ratio. We also report the scaling error, defined as the average difference between the prompted scaling ratio and the actual edited ratio.

\subsection{Quantitative Evaluation}
We show the quantitative comparison with state-of-the-art image editing methods, including diffusion-based editing~\cite{fluxkontext}, unified model~\cite{bagel}, our baseline~\cite{qwenimage} and GPT-Image-1. \cref{tab:quantitative_translation} reports~\name's result on our synthetic evaluation benchmark for three geometric transformation tasks, featuring 100 input samples under each transformation type. Our model outperforms open-sourced baselines among most metrics. We also sample 85 real images from OpenImagesV6~\cite{OpenImages2} for evaluation in \cref{tab:quantitative_real} and results prove our model's effectiveness on real images.

Moreover, a user study is also conducted to further evaluate the editing accuracy and scene coherence. We invite 15 users with 3-year + expertise in multi-modality generation to select a winner among editing candidates from each method and report the winning rate over 30 questions. Results in~\cref{tab:quantitative_translation} demonstrates that~\name~achieves the highest winning rate with a sweeping edge.

\subsection{Qualitative Evaluation}
\begin{figure*}[t]
    \centering
    \includegraphics[width=\linewidth]{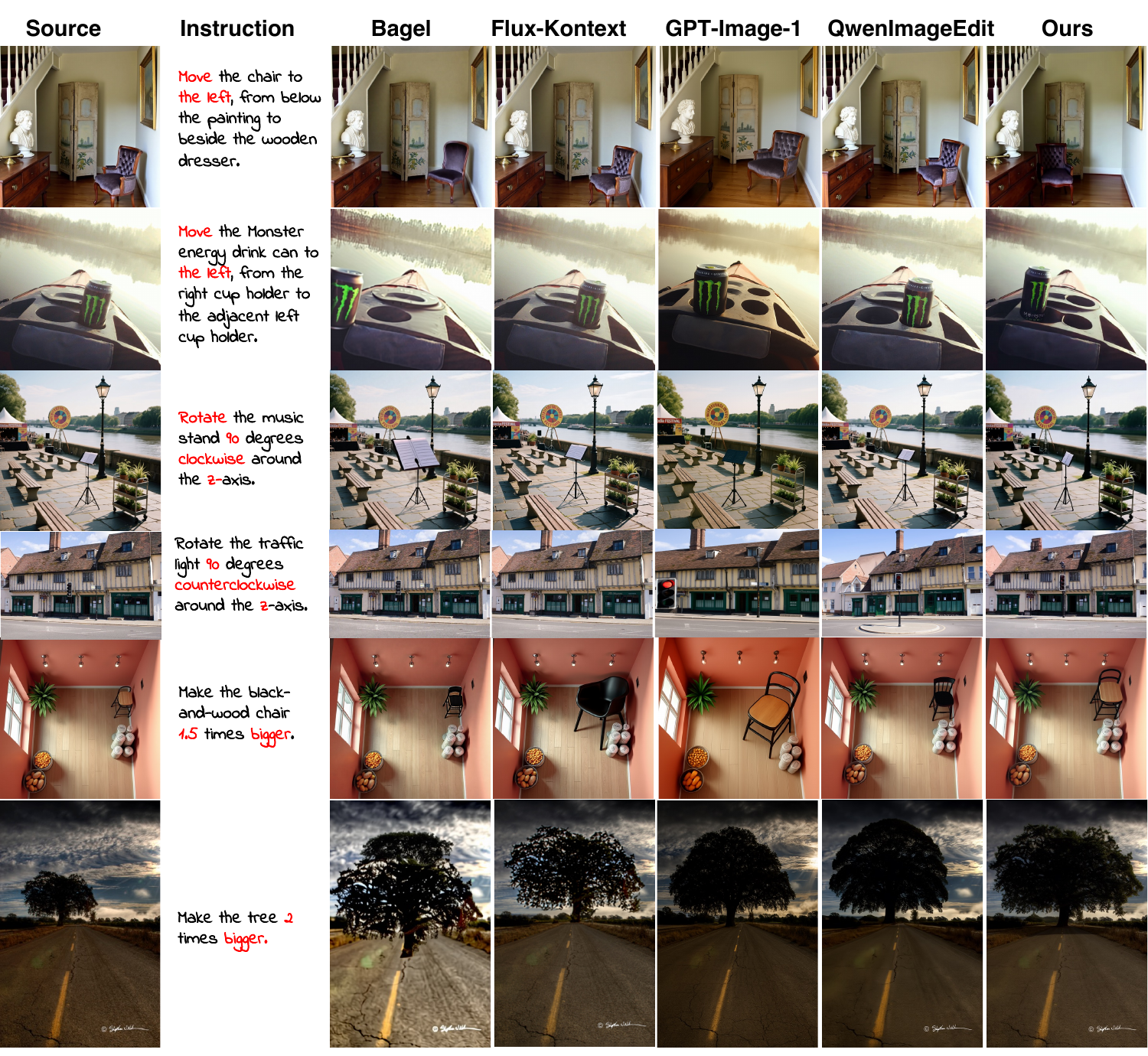}
    \caption{\small \textbf{Qualitative results} on object translation, rotation and resize over state-of-the-art image editing models. For each task, we provide one real image editing result (source from OpenImagesV6~\cite{OpenImages2}) and one synthetic image editing result to showcase the generalization ability of~\name.}
    \label{fig:qualitative}
    \vspace{-4mm}
\end{figure*}

In \cref{fig:qualitative}, we provide multiple qualitative examples including both real and synthetic inputs to showcase~\name’s ability in object transformation compared to other baseline methods. We find that GPT-Image-1 performs transformation edits well, but its result do not preserve the scene details from the reference image, usually exhibits a warmer tone and brighter lighting. Compared to the baseline methods,~\name~executes the instruction more accurately and better preserves the original scene details.

\begin{table}[t]
\small
\centering
\setlength{\tabcolsep}{10pt}
\renewcommand{\arraystretch}{0.8}
\caption{\small \textbf{Ablation study on SFT and RL} under object translation task in terms of translation distance, editing accuracy and image L1 distance.}
\scalebox{0.95}{
\begin{threeparttable}
\begin{tabular}{@{}lcccc@{}}
\toprule
Method & Trans. Dist.$\uparrow$ & Acc.$\uparrow$ & L1$\downarrow$ \\ \midrule
Flux-Kontext & 0.0499 & 4.41\% & 0.5704  \\
Flux-Kontext+SFT & 0.3692 & 38.81\% & 0.5403  \\
Ours & \textbf{0.4798} & \textbf{47.14\%} & \textbf{0.5302}  \\ \midrule
QwenImageEdit & 0.2551 & 32.86\% & 0.5834 \\
QwenImageEdit+SFT & 0.5953 & 67.14\% & 0.2562 \\
Ours & \textbf{0.6667} & \textbf{73.13\%} & \textbf{0.2012}  \\ \midrule
QwenImageEdit+SFT\tnote{†} & 0.2257 & 26.67\% & 0.5928  \\
Ours\tnote{†} & \textbf{0.6507} & \textbf{73.33\%} & \textbf{0.2629}  \\
\bottomrule
\end{tabular}
\begin{tablenotes}[para,flushleft]
\footnotesize
\item[†] Trained with 1/10 of the original training samples.
\end{tablenotes}
\end{threeparttable}}
\vspace{-4pt}
  \label{tab:sftvsrl}
\end{table}

\begin{table}[t]
    \small
    \centering
    \setlength{\tabcolsep}{0.5pt}
    \renewcommand{\arraystretch}{0.9} 
        \caption{\small \textbf{Ablation on active step sampling} under translation task, efficiency measured in seconds(s).}
        \vspace{-2pt}
\scalebox{0.88}{\begin{tabular}{@{}lccccc|cc@{}}
\toprule
Sampling & $\text{NFE}_{\pi_{\theta_\textbf{old}}}$ & $\text{NFE}_{\pi_\theta} $ & {Sample} & {Train} & {Total} & Trans. Dist. & Acc.\\ \cmidrule(lr){0-5} \cmidrule(lr){7-8}
Full      & 10                         & 10                   & 45.78                        & 126.25                       & 172.32    & 0.6602   & 69.12\%             \\
Sliding window & 10                         & 4                    & 45.63                        & 55.69                        & 101.61    &  0.5983 &	67.14\%               \\
Ours          & 4                          & 4                    & \textbf{30.95}                        &\textbf{56.02}                        & \textbf{87.27}  & \textbf{0.6667}                       & \textbf{76.67\%}  \\                
\bottomrule
\end{tabular}}

    \vspace{-8pt}
  \label{tab:efficiency}
\end{table}

\begin{table}[t]
    \small
    \centering
    \setlength{\tabcolsep}{1pt}
    \renewcommand{\arraystretch}{1.0} 
    \caption{\small \textbf{Background ID Consistency} of three object transformation tasks in terms of image-level CLIP and L1 distance. The best and second best results are highlighted with \textbf{bold} and \underline{underline}.} 
\scalebox{0.85}{
\begin{tabular}{@{}lcccccc@{}}
\toprule
\multicolumn{1}{c}{\multirow{2}{*}{Method}}            & \multicolumn{2}{c}{Translation}                & \multicolumn{2}{c}{Rotation}                                                              & \multicolumn{2}{c}{Resize}                                                                \\ \cmidrule(l){2-7} 
\multicolumn{1}{c}{}                                   & CLIP-im~$\uparrow$ & L1~$\downarrow$ & CLIP-im~$\uparrow$ & \begin{tabular}[c]{@{}c@{}}L1~$\downarrow$\end{tabular} & CLIP-im~$\uparrow$ & \begin{tabular}[c]{@{}c@{}}L1~$\downarrow$\end{tabular} \\ \midrule
GPT-Image-1~\cite{openai2024gptimage1}                                           & 0.9318                  & 0.4351               & 0.9453                  & \underline{0.4136}                                                    & 0.9405                  & 0.4278                                                          \\
Flux-Kontext~\cite{fluxkontext} & 0.9660                  & 0.5704               & \textbf{0.9821}         & 0.5390                                                          & \underline{0.9739}            & 0.5145                                                          \\
Bagel~\cite{bagel}             & 0.9570                  & \underline{0.3308}         & 0.9356                  & 0.4138                                                          & \textbf{0.9790}         & \textbf{0.3027}                                                 \\
QwenImageEdit~\cite{qwenimage} & \textbf{0.9728}         & 0.5834               & \underline{0.9777}            & 0.6000                                                          & 0.9724                  & 0.5869                                                          \\
Ours                                                   & \underline{0.9699}            & \textbf{0.2012}      & 0.9717                  & \textbf{0.2774}                                                 & 0.9705                  & \underline{0.3841}                                                    \\ \bottomrule
\end{tabular}
}

    \vspace{-10pt}
  \label{tab:consistency}
\end{table}

\begin{figure*}[ht]
    \centering
    \begin{subfigure}[t]{0.31\linewidth}
        \centering
        \includegraphics[width=\linewidth]{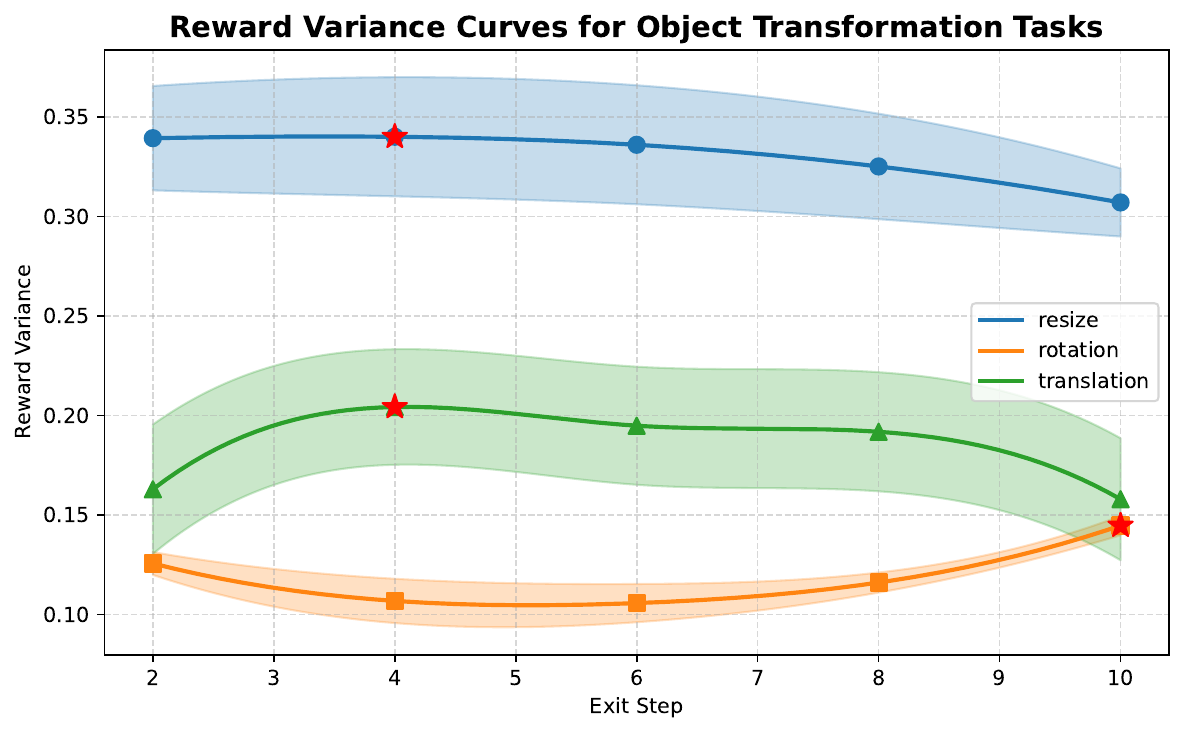}
        \caption{\textbf{Off-policy reward variance distribution} for translation, rotation and resizing.}
        \label{fig:variance}
    \end{subfigure}
    \hfill
    %
    \begin{subfigure}[t]{0.39\linewidth}
        \centering
        \includegraphics[width=\linewidth]{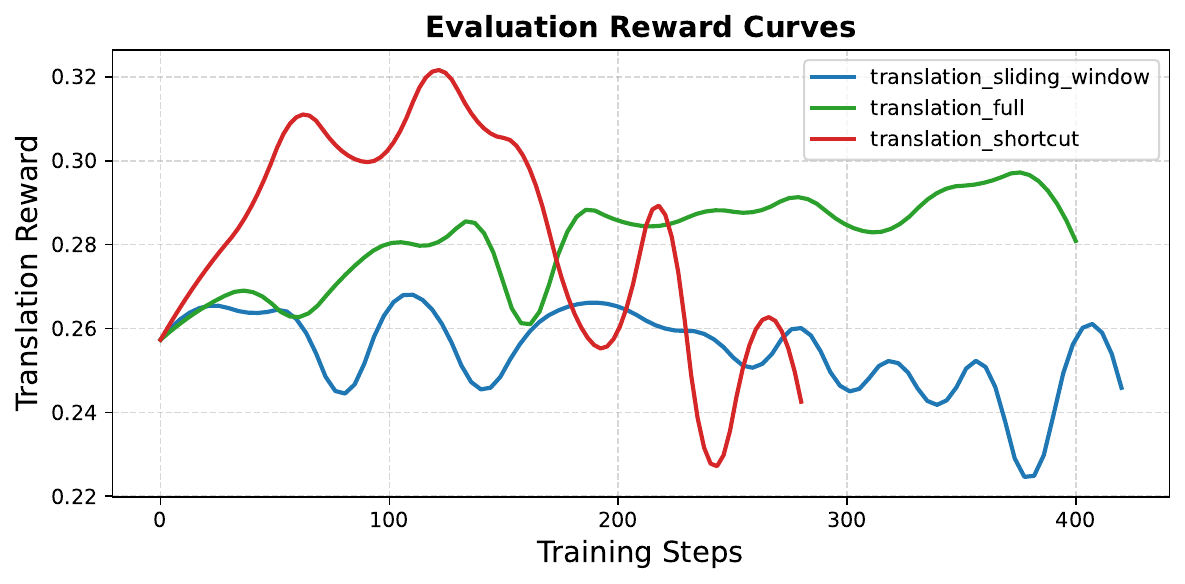}
        \caption{Reward curves for \textbf{different sampling strategies} under the translation task.}
        \label{fig:evalcurve}
    \end{subfigure}
     \hfill
    %
    \begin{subfigure}[t]{0.27\linewidth}
        \centering
        \includegraphics[width=\linewidth]{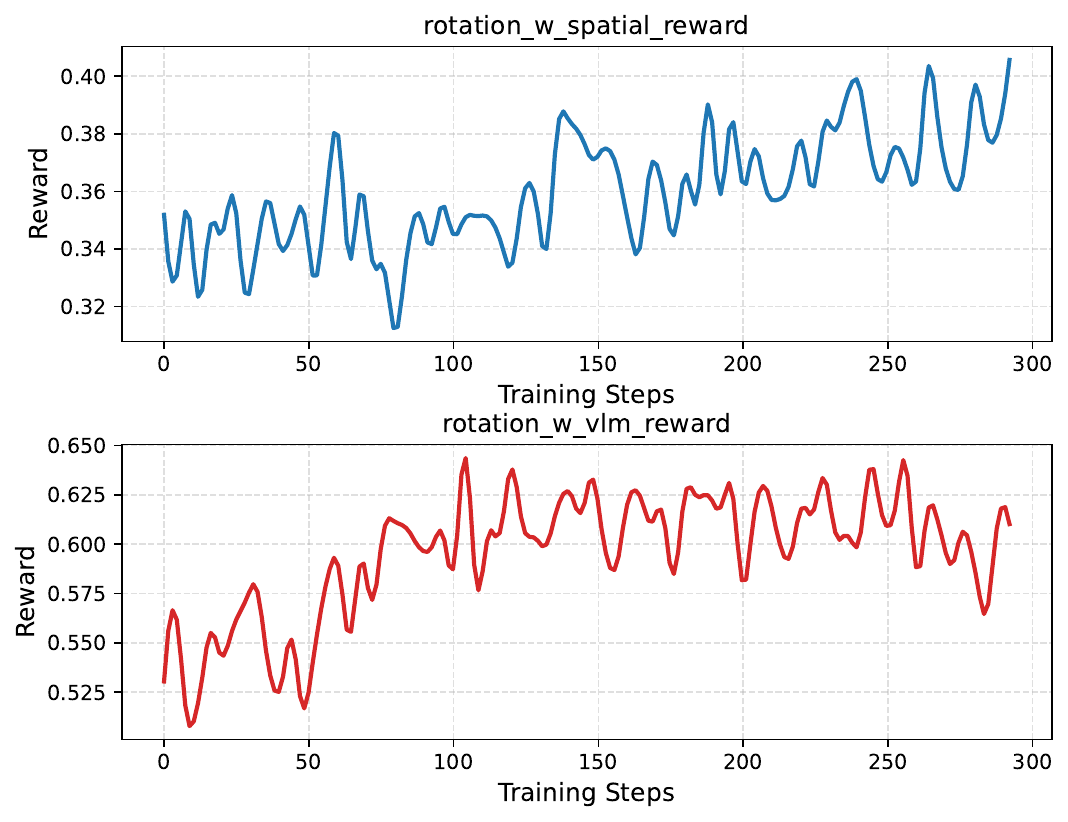}
        \caption{\textbf{Ablative comparison} of VLM reward vs. spatial reward under the rotation task.}
        \label{fig:ablation_reward}
    \end{subfigure}

    \vspace{-2mm}
    \caption{\textbf{Reward behavior across tasks}: (a) reward variance distribution, (b) GRPO sampling strategies, and (c) reward model ablations.}
    \label{fig:three_reward_figs}
    \vspace{-2mm}
\end{figure*}
\subsection{Ablation Studies}

\noindent\textbf{SFT vs RL.}
\cref{tab:sftvsrl} compares how SFT and RL contribute to spatial manipulation performance. We evaluate both approaches on two backbone models: the widely used editing backbone Flux-Kontext~\cite{fluxkontext} and the more advanced QwenImageEdit model~\cite{qwenimage}, which integrates a VLM with a diffusion generator. To ensure a fair comparison, all experiments are trained solely on object translation data.

With a reasonably large training set (800 images), SFT provides a strong initial boost in editing accuracy. However, once SFT converges, applying RL on top of the SFT checkpoint further increases both the average translation distance and the final accuracy for both backbones.

To further assess the data efficiency of RL, we reduce the training data to one tenth of the original size ($\sim$80 annotated pairs and $\sim$400 input-only pairs). As shown in the third block of \cref{tab:sftvsrl}, SFT fails to achieve meaningful gains under such limited data, whereas RL built upon the small-data SFT checkpoint still reaches performance comparable to the full-data setting. This demonstrates that even with very limited input data, a well-designed reward function allows RL to effectively learn spatial manipulation behaviors and achieve competitive editing accuracy.

\noindent\textbf{Background Scene Preservation.} A good editing model should preserve the background identity while manipulating the foreground. We evaluate scene consistency using the image-level L1 distance between the input and edited images. As shown in \cref{tab:consistency}, our method achieves an L1 distance comparable to other state-of-the-art editing models, indicating that it does not compromise scene consistency during training.

The CLIP metric is relatively coarse and not sensitive to subtle background changes; only large deviations meaningfully indicate identity loss. This is also evident from our qualitative results, where methods with similar CLIP-im scores can still show almost the same level of background preservation.
In contrast, GPT-Image-1 exhibits noticeably lower CLIP similarity than the other methods. This is because GPT-Image-1 tends to heavily refine or alter the entire image, which is also evident in our qualitative results.

\noindent\textbf{Off-policy Step Evaluation Results.}
We present the off-policy evaluation results for the three object transformation tasks. As shown in \cref{fig:variance}, the maximum reward variance occurs at step 4 (out of 10) for both translation and resizing, while for rotation it appears at step 10. These different exit steps demonstrate that in flow-based diffusion, injecting the same Gaussian noise at different steps leads to varying reward sensitivity and exploration space across tasks.

\noindent\textbf{Ablation on Active Step Sampling.}
We further study the time efficiency of our method. \cref{tab:efficiency} reports the sampling and optimization time for a batch of 16 rollouts. Based on the off-policy step analysis, we adopt step 4 as the exit step for the translation task, resulting in only 4 sampled steps per rollout. We observe that our method reduces the overall iteration time by 14\% compared to the sliding-window baseline (translation–sliding window), and by 49\% compared to full sampling (translation–full). At the same time, despite using fewer sampled steps, our shortcut strategy still achieves higher editing correctness, outperforming both baselines in terms of Translation Distance and Accuracy.

We also plot the evaluation reward curves in \cref{fig:evalcurve} for both baselines and our method. Interestingly, taking shortcuts not only reduces computation but also accelerates convergence, further suggesting that focusing on the steps that provide larger geometric transformation space is beneficial for exploration and overall performance.

\noindent\textbf{Ablation on Reward Model.}
We conduct an ablation comparing a VLM-based reward model with our spatial reward derived from a specialist model (Orient-Anything~\cite{orientanything} + rule-based evaluation). Editing quality is assessed using Qwen2.5-VL-Instruct~\cite{Qwen2.5-VL} to measure alignment between the edited image and the prompt. As shown in \cref{fig:ablation_reward}, both reward models improve evaluation reward during GRPO training. However, the VLM-based reward often produces overly optimistic and unstable scores, yielding a rotation error of 0.3294 and an accuracy of 11.63\% according to evaluation. In contrast, our spatial reward leads to a lower error of 0.2861 and a much higher accuracy of 29.55\%, indicating that it provides a more reliable supervision signal for GRPO-based spatial manipulation.

\section{Conclusion and Discussion}
In this work, we presented \name, the first reinforcement learning framework for text-guided geometric object transformation in scene editing. Built on GRPO-style training, \name enables data-efficient learning with reduced reliance on expensive paired data and uses spatially grounded rewards to disentangle objects from the scene and improve fine-grained transformations. We further enhance computational efficiency through off-policy step evaluation and active step sampling, prioritizing informative steps with higher reward variance and larger transformation space. Experiments show that~\name~outperforms existing open- and closed-source models in spatial accuracy and visual coherence. 
Our RL-based recipe can be extended to other generative frameworks (e.g., GANs, autoregressive models) for verifiable, controllable visual generation, which we leave for future work.

{
    \small
    \bibliographystyle{ieeenat_fullname}
    \bibliography{main}

@article{flowgrpo,
  title={Flow-grpo: Training flow matching models via online rl},
  author={Liu, Jie and Liu, Gongye and Liang, Jiajun and Li, Yangguang and Liu, Jiaheng and Wang, Xintao and Wan, Pengfei and Zhang, Di and Ouyang, Wanli},
  journal={arXiv preprint arXiv:2505.05470},
  year={2025}
}

@article{dancegrpo,
  title={DanceGRPO: Unleashing GRPO on Visual Generation},
  author={Xue, Zeyue and Wu, Jie and Gao, Yu and Kong, Fangyuan and Zhu, Lingting and Chen, Mengzhao and Liu, Zhiheng and Liu, Wei and Guo, Qiushan and Huang, Weilin and others},
  journal={arXiv preprint arXiv:2505.07818},
  year={2025}
}

@article{mixgrpo,
  title={Mixgrpo: Unlocking flow-based grpo efficiency with mixed ode-sde},
  author={Li, Junzhe and Cui, Yutao and Huang, Tao and Ma, Yinping and Fan, Chun and Yang, Miles and Zhong, Zhao},
  journal={arXiv preprint arXiv:2507.21802},
  year={2025}
}

@article{g2rpo,
  title={G2RPO: Granular GRPO for Precise Reward in Flow Models},
  author={Zhou, Yujie and Ling, Pengyang and Bu, Jiazi and Wang, Yibin and Zang, Yuhang and Wang, Jiaqi and Niu, Li and Zhai, Guangtao},
  journal={arXiv preprint arXiv:2510.01982},
  year={2025}
}

@article{bagel,
  title   = {Emerging Properties in Unified Multimodal Pretraining},
  author  = {Deng, Chaorui and Zhu, Deyao and Li, Kunchang and Gou, Chenhui and Li, Feng and Wang, Zeyu and Zhong, Shu and Yu, Weihao and Nie, Xiaonan and Song, Ziang and Shi, Guang and Fan, Haoqi},
  journal = {arXiv preprint arXiv:2505.14683},
  year    = {2025}
}

@misc{qwenimage,
      title={Qwen-Image Technical Report}, 
      author={Chenfei Wu and Jiahao Li and Jingren Zhou and Junyang Lin and Kaiyuan Gao and Kun Yan and Sheng-ming Yin and Shuai Bai and Xiao Xu and Yilei Chen and Yuxiang Chen and Zecheng Tang and Zekai Zhang and Zhengyi Wang and An Yang and Bowen Yu and Chen Cheng and Dayiheng Liu and Deqing Li and Hang Zhang and Hao Meng and Hu Wei and Jingyuan Ni and Kai Chen and Kuan Cao and Liang Peng and Lin Qu and Minggang Wu and Peng Wang and Shuting Yu and Tingkun Wen and Wensen Feng and Xiaoxiao Xu and Yi Wang and Yichang Zhang and Yongqiang Zhu and Yujia Wu and Yuxuan Cai and Zenan Liu},
      year={2025},
      eprint={2508.02324},
      archivePrefix={arXiv},
      primaryClass={cs.CV},
      url={https://arxiv.org/abs/2508.02324}, 
}

@misc{fluxkontext,
      title={FLUX.1 Kontext: Flow Matching for In-Context Image Generation and Editing in Latent Space}, 
      author={Black Forest Labs and Stephen Batifol and Andreas Blattmann and Frederic Boesel and Saksham Consul and Cyril Diagne and Tim Dockhorn and Jack English and Zion English and Patrick Esser and Sumith Kulal and Kyle Lacey and Yam Levi and Cheng Li and Dominik Lorenz and Jonas Müller and Dustin Podell and Robin Rombach and Harry Saini and Axel Sauer and Luke Smith},
      year={2025},
      eprint={2506.15742},
      archivePrefix={arXiv},
      primaryClass={cs.GR},
      url={https://arxiv.org/abs/2506.15742},
}

@article{branchgrpo,
  title={Branchgrpo: Stable and efficient grpo with structured branching in diffusion models},
  author={Li, Yuming and Wang, Yikai and Zhu, Yuying and Zhao, Zhongyu and Lu, Ming and She, Qi and Zhang, Shanghang},
  journal={arXiv preprint arXiv:2509.06040},
  year={2025}
}

@article{geneval,
  title={Geneval: An object-focused framework for evaluating text-to-image alignment},
  author={Ghosh, Dhruba and Hajishirzi, Hannaneh and Schmidt, Ludwig},
  journal={Advances in Neural Information Processing Systems},
  volume={36},
  pages={52132--52152},
  year={2023}
}

@article{orientanything,
  title={Orient anything: Learning robust object orientation estimation from rendering 3d models},
  author={Wang, Zehan and Zhang, Ziang and Pang, Tianyu and Du, Chao and Zhao, Hengshuang and Zhao, Zhou},
  journal={arXiv preprint arXiv:2412.18605},
  year={2024}
}

@inproceedings{stablediffusion,
  title={High-resolution image synthesis with latent diffusion models},
  author={Rombach, Robin and Blattmann, Andreas and Lorenz, Dominik and Esser, Patrick and Ommer, Bj{\"o}rn},
  booktitle={Proceedings of the IEEE/CVF conference on computer vision and pattern recognition},
  pages={10684--10695},
  year={2022}
}

@inproceedings{ramesh2021zero,
  title={Zero-shot text-to-image generation},
  author={Ramesh, Aditya and Pavlov, Mikhail and Goh, Gabriel and Gray, Scott and Voss, Chelsea and Radford, Alec and Chen, Mark and Sutskever, Ilya},
  booktitle={International conference on machine learning},
  pages={8821--8831},
  year={2021},
  organization={Pmlr}
}

@misc{blackforestlabs2023flux,
  author       = {Black Forest Labs},
  title        = {FLUX},
  year         = {2023},
  howpublished = {\url{https://github.com/black-forest-labs/flux}},
}

@article{metaqueries,
  title={Transfer between modalities with metaqueries},
  author={Pan, Xichen and Shukla, Satya Narayan and Singh, Aashu and Zhao, Zhuokai and Mishra, Shlok Kumar and Wang, Jialiang and Xu, Zhiyang and Chen, Jiuhai and Li, Kunpeng and Juefei-Xu, Felix and others},
  journal={arXiv preprint arXiv:2504.06256},
  year={2025}
}

@article{seedx,
  title={Seed-x: Multimodal models with unified multi-granularity comprehension and generation},
  author={Ge, Yuying and Zhao, Sijie and Zhu, Jinguo and Ge, Yixiao and Yi, Kun and Song, Lin and Li, Chen and Ding, Xiaohan and Shan, Ying},
  journal={arXiv preprint arXiv:2404.14396},
  year={2024}
}

@inproceedings{emu2,
  title={Generative multimodal models are in-context learners},
  author={Sun, Quan and Cui, Yufeng and Zhang, Xiaosong and Zhang, Fan and Yu, Qiying and Wang, Yueze and Rao, Yongming and Liu, Jingjing and Huang, Tiejun and Wang, Xinlong},
  booktitle={Proceedings of the IEEE/CVF Conference on Computer Vision and Pattern Recognition},
  pages={14398--14409},
  year={2024}
}

@article{mot,
  title={Mixture-of-transformers: A sparse and scalable architecture for multi-modal foundation models},
  author={Liang, Weixin and Yu, Lili and Luo, Liang and Iyer, Srinivasan and Dong, Ning and Zhou, Chunting and Ghosh, Gargi and Lewis, Mike and Yih, Wen-tau and Zettlemoyer, Luke and others},
  journal={arXiv preprint arXiv:2411.04996},
  year={2024}
}

@inproceedings{rwr,
  title={Reinforcement learning by reward-weighted regression for operational space control},
  author={Peters, Jan and Schaal, Stefan},
  booktitle={Proceedings of the 24th international conference on Machine learning},
  pages={745--750},
  year={2007}
}

@article{ddpo,
  title={Training diffusion models with reinforcement learning},
  author={Black, Kevin and Janner, Michael and Du, Yilun and Kostrikov, Ilya and Levine, Sergey},
  journal={arXiv preprint arXiv:2305.13301},
  year={2023}
}

@article{editscore,
  title={EditScore: Unlocking Online RL for Image Editing via High-Fidelity Reward Modeling},
  author={Luo, Xin and Wang, Jiahao and Wu, Chenyuan and Xiao, Shitao and Jiang, Xiyan and Lian, Defu and Zhang, Jiajun and Liu, Dong and others},
  journal={arXiv preprint arXiv:2509.23909},
  year={2025}
}

@article{3ditscene,
  title={3ditscene: Editing any scene via language-guided disentangled gaussian splatting},
  author={Zhang, Qihang and Xu, Yinghao and Wang, Chaoyang and Lee, Hsin-Ying and Wetzstein, Gordon and Zhou, Bolei and Yang, Ceyuan},
  journal={arXiv preprint arXiv:2405.18424},
  year={2024}
}

@article{wang2025beyond,
  title={Beyond the 80/20 rule: High-entropy minority tokens drive effective reinforcement learning for llm reasoning},
  author={Wang, Shenzhi and Yu, Le and Gao, Chang and Zheng, Chujie and Liu, Shixuan and Lu, Rui and Dang, Kai and Chen, Xionghui and Yang, Jianxin and Zhang, Zhenru and others},
  journal={arXiv preprint arXiv:2506.01939},
  year={2025}
}

@article{gao2025seedream,
  title={Seedream 3.0 technical report},
  author={Gao, Yu and Gong, Lixue and Guo, Qiushan and Hou, Xiaoxia and Lai, Zhichao and Li, Fanshi and Li, Liang and Lian, Xiaochen and Liao, Chao and Liu, Liyang and others},
  journal={arXiv preprint arXiv:2504.11346},
  year={2025}
}

@inproceedings{sam,
  title={Segment anything},
  author={Kirillov, Alexander and Mintun, Eric and Ravi, Nikhila and Mao, Hanzi and Rolland, Chloe and Gustafson, Laura and Xiao, Tete and Whitehead, Spencer and Berg, Alexander C and Lo, Wan-Yen and others},
  booktitle={Proceedings of the IEEE/CVF international conference on computer vision},
  pages={4015--4026},
  year={2023}
}

@inproceedings{groundingdino,
  title={Grounding dino: Marrying dino with grounded pre-training for open-set object detection},
  author={Liu, Shilong and Zeng, Zhaoyang and Ren, Tianhe and Li, Feng and Zhang, Hao and Yang, Jie and Jiang, Qing and Li, Chunyuan and Yang, Jianwei and Su, Hang and others},
  booktitle={European conference on computer vision},
  pages={38--55},
  year={2024},
  organization={Springer}
}

@misc{openai2024gptimage1,
  author       = {OpenAI},
  title        = {GPT-Image-1: OpenAI’s Image Generation Model},
  year         = {2024},
  howpublished = {\url{https://platform.openai.com/docs/models/gpt-image-1}},
  note         = {Accessed: 2025-11-11}
}

@inproceedings{depthanything,
  title={Depth anything: Unleashing the power of large-scale unlabeled data},
  author={Yang, Lihe and Kang, Bingyi and Huang, Zilong and Xu, Xiaogang and Feng, Jiashi and Zhao, Hengshuang},
  booktitle={Proceedings of the IEEE/CVF conference on computer vision and pattern recognition},
  pages={10371--10381},
  year={2024}
}

@inproceedings{wang2025moge,
  title={Moge: Unlocking accurate monocular geometry estimation for open-domain images with optimal training supervision},
  author={Wang, Ruicheng and Xu, Sicheng and Dai, Cassie and Xiang, Jianfeng and Deng, Yu and Tong, Xin and Yang, Jiaolong},
  booktitle={Proceedings of the Computer Vision and Pattern Recognition Conference},
  pages={5261--5271},
  year={2025}
}

@article{gallagher2015happened,
  title={What happened to my dog in that network: Unraveling top-down generators in convolutional neural networks},
  author={Gallagher, Patrick W and Tang, Shuai and Tu, Zhuowen},
  journal={arXiv preprint arXiv:1511.07125},
  year={2015}
}

@article{reed2015deep,
  title={Deep visual analogy-making},
  author={Reed, Scott E and Zhang, Yi and Zhang, Yuting and Lee, Honglak},
  journal={Advances in neural information processing systems},
  volume={28},
  year={2015}
}

@inproceedings{dragyourgan,
  title={Drag your gan: Interactive point-based manipulation on the generative image manifold},
  author={Pan, Xingang and Tewari, Ayush and Leimk{\"u}hler, Thomas and Liu, Lingjie and Meka, Abhimitra and Theobalt, Christian},
  booktitle={ACM SIGGRAPH 2023 conference proceedings},
  pages={1--11},
  year={2023}
}

@inproceedings{3dfixup,
  title={3d-fixup: Advancing photo editing with 3d priors},
  author={Cheng, Yen-Chi and Singh, Krishna Kumar and Yoon, Jae Shin and Schwing, Alexander and Gui, Liang-Yan and Gadelha, Matheus and Guerrero, Paul and Zhao, Nanxuan},
  booktitle={Proceedings of the Special Interest Group on Computer Graphics and Interactive Techniques Conference Conference Papers},
  pages={1--10},
  year={2025}
}

@article{mou2023dragondiffusion,
  title={Dragondiffusion: Enabling drag-style manipulation on diffusion models},
  author={Mou, Chong and Wang, Xintao and Song, Jiechong and Shan, Ying and Zhang, Jian},
  journal={arXiv preprint arXiv:2307.02421},
  year={2023}
}

@article{li2024omnidrag,
  title={OmniDrag: Enabling Motion Control for Omnidirectional Image-to-Video Generation},
  author={Li, Weiqi and Zhao, Shijie and Mou, Chong and Sheng, Xuhan and Zhang, Zhenyu and Wang, Qian and Li, Junlin and Zhang, Li and Zhang, Jian},
  journal={arXiv preprint arXiv:2412.09623},
  year={2024}
}

@inproceedings{lu2024regiondrag,
  title={Regiondrag: Fast region-based image editing with diffusion models},
  author={Lu, Jingyi and Li, Xinghui and Han, Kai},
  booktitle={European Conference on Computer Vision},
  pages={231--246},
  year={2024},
  organization={Springer}
}

@inproceedings{imagesculpting,
  title={Image sculpting: Precise object editing with 3d geometry control},
  author={Yenphraphai, Jiraphon and Pan, Xichen and Liu, Sainan and Panozzo, Daniele and Xie, Saining},
  booktitle={Proceedings of the IEEE/CVF Conference on Computer Vision and Pattern Recognition},
  pages={4241--4251},
  year={2024}
}

@inproceedings{dragyournoise,
  title={Drag your noise: Interactive point-based editing via diffusion semantic propagation},
  author={Liu, Haofeng and Xu, Chenshu and Yang, Yifei and Zeng, Lihua and He, Shengfeng},
  booktitle={Proceedings of the IEEE/CVF conference on computer vision and pattern recognition},
  pages={6743--6752},
  year={2024}
}

@inproceedings{shi2024dragdiffusion,
  title={Dragdiffusion: Harnessing diffusion models for interactive point-based image editing},
  author={Shi, Yujun and Xue, Chuhui and Liew, Jun Hao and Pan, Jiachun and Yan, Hanshu and Zhang, Wenqing and Tan, Vincent YF and Bai, Song},
  booktitle={Proceedings of the IEEE/CVF Conference on Computer Vision and Pattern Recognition},
  pages={8839--8849},
  year={2024}
}

@inproceedings{shin2024instantdrag,
  title={Instantdrag: Improving interactivity in drag-based image editing},
  author={Shin, Joonghyuk and Choi, Daehyeon and Park, Jaesik},
  booktitle={SIGGRAPH Asia 2024 Conference Papers},
  pages={1--10},
  year={2024}
}

@article{Qwen2.5-VL,
  title={Qwen2.5-VL Technical Report},
  author={Bai, Shuai and Chen, Keqin and Liu, Xuejing and Wang, Jialin and Ge, Wenbin and Song, Sibo and Dang, Kai and Wang, Peng and Wang, Shijie and Tang, Jun and Zhong, Humen and Zhu, Yuanzhi and Yang, Mingkun and Li, Zhaohai and Wan, Jianqiang and Wang, Pengfei and Ding, Wei and Fu, Zheren and Xu, Yiheng and Ye, Jiabo and Zhang, Xi and Xie, Tianbao and Cheng, Zesen and Zhang, Hang and Yang, Zhibo and Xu, Haiyang and Lin, Junyang},
  journal={arXiv preprint arXiv:2502.13923},
  year={2025}
}

@article{OpenImages2,
  title={OpenImages: A public dataset for large-scale multi-label and multi-class image classification.},
  author={Krasin, Ivan and Duerig, Tom and Alldrin, Neil and Ferrari, Vittorio and Abu-El-Haija, Sami and Kuznetsova, Alina and Rom, Hassan and Uijlings, Jasper and Popov, Stefan and Kamali, Shahab and Malloci, Matteo and Pont-Tuset, Jordi and Veit, Andreas and Belongie, Serge and Gomes, Victor and Gupta, Abhinav and Sun, Chen and Chechik, Gal and Cai, David and Feng, Zheyun and Narayanan, Dhyanesh and Murphy, Kevin},
  journal={Dataset available from https://storage.googleapis.com/openimages/web/index.html},
  year={2017}
}

@article{li2024instructrl4pix,
  title={Instructrl4pix: Training diffusion for image editing by reinforcement learning},
  author={Li, Tiancheng and Liu, Jinxiu and Chen, Huajun and Liu, Qi},
  journal={arXiv preprint arXiv:2406.09973},
  year={2024}
}

@article{li2025uniworld,
  title={Uniworld-V2: Reinforce Image Editing with Diffusion Negative-aware Finetuning and MLLM Implicit Feedback},
  author={Li, Zongjian and Liu, Zheyuan and Zhang, Qihui and Lin, Bin and Yuan, Shenghai and Yan, Zhiyuan and Ye, Yang and Yu, Wangbo and Niu, Yuwei and Yuan, Li},
  journal={arXiv preprint arXiv:2510.16888},
  year={2025}
}

@article{he2025tempflow,
  title={Tempflow-grpo: When timing matters for grpo in flow models},
  author={He, Xiaoxuan and Fu, Siming and Zhao, Yuke and Li, Wanli and Yang, Jian and Yin, Dacheng and Rao, Fengyun and Zhang, Bo},
  journal={arXiv preprint arXiv:2508.04324},
  year={2025}
}
}

\end{document}